\documentclass[runningheads]{llncs}
\usepackage{graphicx}
\usepackage{comment}
\usepackage{amsmath,amssymb} 
\usepackage{color}
\usepackage{booktabs}

\newcommand{\secref}[1]{Section~\ref{#1}}
\newcommand{\figref}[1]{Figure~\ref{#1}}
\newcommand{\tabref}[1]{Table~\ref{#1}}

\newcommand{\rot}[1]{\rotatebox[origin=c]{90}{#1}}

\begin{document}
\pagestyle{headings}
\mainmatter
\def\ECCVSubNumber{100}  

\title{Robust Vision Challenge 2020 -\\$1^{\text{st}}$ Place Report for Panoptic Segmentation} 


\titlerunning{1st Place RVC 2020 - Panoptic Segmentation}
%
\author{Rohit Mohan \and
Abhinav Valada}
\authorrunning{R. Mohan and A. Valada}
%
\institute{University of Freiburg\\ \email{\{mohan, valada\}@cs.uni-freiburg.de}}
\maketitle

\begin{abstract}
In this technical report, we present key details of our winning panoptic segmentation architecture ‘EffPS\_b1bs4\_RVC’. Our network is a lightweight version of our state-of-the-art EfficientPS architecture that consists of our proposed shared backbone with a modified EfficientNet-B5 model as the encoder, followed by the 2-way FPN to learn semantically rich multi-scale features. It consists of two task-specific heads, a modified Mask R-CNN instance head and our novel semantic segmentation head that processes features of different scales with specialized modules for coherent feature refinement. Finally, our proposed panoptic fusion module adaptively fuses logits from each of the heads to yield the panoptic segmentation output. The Robust Vision Challenge 2020 benchmarking results show that our model is ranked \#1 on Microsoft COCO, VIPER and WildDash, and is ranked \#2 on Cityscapes and Mapillary Vistas, thereby achieving the overall rank \#1 for the panoptic segmentation task.\looseness=-1

\keywords{Panoptic Segmentation \and Semantic Segmentation \and Instance Segmentation \and Scene Understanding}
\end{abstract}

\section{Introduction}

Holistic scene understanding plays an critical role in the competent functioning of an autonomous robot. It enables the ability to perceive and reason which allows robots to interact with their surroundings. Broadly, the components of a scene can be classified into ‘stuff’ and ‘thing’ categories, where ‘stuff’ classes are defined as uncountable amorphous regions such as road, sidewalk, and buildings, while ‘thing’ classes are defined as countable objects such as people, cars, and cyclists. The panoptic segmentation task~\cite{kirillov2019panoptic} aims to jointly predict ‘stuff’ and ‘thing’ classes using a single coherent Convolutional Neural Network (CNN) architecture. More precisely, the network assigns a semantic ID to a pixel of an image if it belongs to ‘stuff’ classes and it assigns a class label as well as an instance ID if the pixel belongs to a 'thing' class. We address the panoptic segmentation task using our recently proposed Efficient Panoptic Segmentation (EfficientPS)~\cite{mohan2020efficientps} architecture that incorporates our proposed efficient shared backbone with our new feature aligning semantic head, a new variant of Mask R-CNN~\cite{he2017mask} as the instance head, and our novel adaptive panoptic fusion module. 

We benchmark our model on the Robust Vision Challenge (RVC) datasets. The goal of RVC is to encourage the development of vision systems that are robust and that can consequently perform well on a variety of datasets with different characteristics. The RVC panoptic segmentation challenge requires participants to train a single model over a joint dataset comprising of six different datasets and to benchmark individually on each corresponding benchmark server. The models are then ranked using the Schulze Proportional Ranking (PR) method~\cite{schulze2011new}. To address this challenge, we employ a lightweight version of our EfficientPS architecture which we refer to as ‘EffPS\_b1bs4\_RVC’ and it achieves rank \#1 in RVC 2020.

\section{Methodology}

\begin{figure*}[t]
\centering
\includegraphics[width=\linewidth]{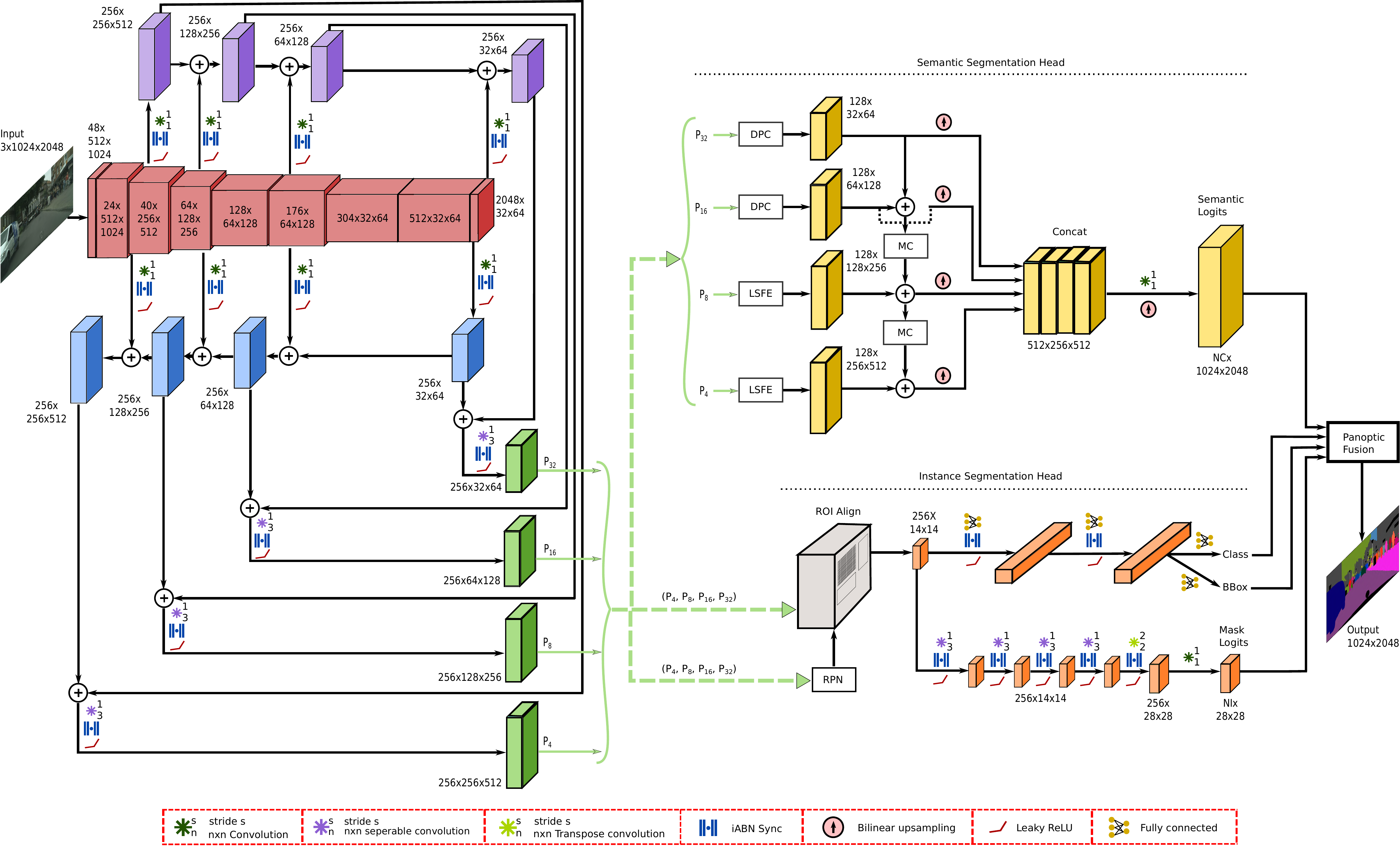}
\caption{Illustration of our EfficientPS~\cite{mohan2020efficientps} architecture comprising of an encoder with the 2-way FPN that constitute the shared backbone, parallel semantic and instance segmentation heads, followed by the adaptive panoptic fusion module that yields the final panoptic segmentation output.}
\label{fig:epsnet_arch}
\end{figure*}

For this challenge, we use our EfficeintPS architecture with minor modifications due to resource and time constraints. We use a modified version of EfficientNet-B5~\cite{tan2019efficientnet} as the encoder in EfficientPS. While, for RVC we use EfficientNet-B1 due to memory limitations. Additionally, EfficientPS employs depthwise separable convolutions in the Mask R-CNN instance head but we use standard convolutions in the RVC model to obtain a minor increase in performance over a reduction in the number of parameters due to the weaker EfficientNet-B1 encoder that we employ.\looseness=-1

For the sake of completeness, we briefly discuss the topology of the EfficientPS architecture shown in \figref{fig:epsnet_arch}. It consists of four components: a shared backbone, instance head, semantic head, and the panoptic fusion module. The shared backbone comprises of a modified EfficientNet-B5 and our novel 2-way FPN architecture. We remove the Squeeze-and Excitation(SE)~\cite{hu2018squeeze} connections of the standard EfficientNet-B5 and replace the batch normalization layers~\cite{ioffe2015batch} with synchronized Inplace Activated Batch Normalization (iABNsync)~\cite{rota2018place}. To overcome the problem of unidirectional flow of information, we add a parallel bottom-up branch to the standard FPN architecture followed by the summation of each branch at corresponding scales to attain the proposed 2-way FPN. The left half of \figref{fig:epsnet_arch} depicts the backbone network.

The instance segmentation head of the EfficientPS architecture comprises of Mask R-CNN with standard convolutions, batch normalization, and ReLU activations layers replaced with depthwise separable convolution, iABN sync, and Leaky ReLU respectively. The novel semantic head uses three different modules to effectively capture features of different scales and to subsequently correlate them to avoid the mismatch problem. Therefore, we employ Dense Prediction Cells(DPC)~\cite{chen2018searching} on small-scale features and our LSFE module~\cite{mohan2020efficientps} on large-scale features. We then align the different scaled features using the Mismatch Correction module~\cite{mohan2020efficientps} before the segmentation logits are obtained as shown in \figref{fig:epsnet_arch}. Finally, the panoptic fusion module adaptively fuses logits from the semantic and instance heads based on their mask confidences. The fusion is performed according to equation \eqref{eq:os} which enables selective attenuation or amplification of fused logit scores to integrate instance-specific ‘thing’ classes with ‘stuff’ classes as
\begin{equation}\label{eq:os}
FL = (\sigma(ML_A) + \sigma(ML_B)) \odot (ML_A + ML_B),
\end{equation}
where $ML_A$ and $ML_B$ are logits from the semantic and instance heads respectively.

\section{Training}

In this section, we first describe the datasets in \secref{sec:datasets} and then present the data augmentation strategies that we employ to mitigate problems induced by combining all the dataset in \secref{sec:augmentation}. Finally, we detail the training protocol that we employ in \secref{sec:protocol}.

We use the PyTorch~\cite{paszke2019pytorch} deep learning library for implementing our architecture and we trained our model on a system with an Intel Xenon@2.20GHz processor with NVIDIA TITAN RTX GPUs.  

\subsection{Datasets}
\label{sec:datasets}

We combine six different panoptic segmentation benchmark datasets to form the joint training set. Namely, we use Microsoft COCO~\cite{lin2014microsoft}, Cityscapes~\cite{cordts2016cityscapes}, Mapillary Vistas Dataset~\cite{neuhold2017mapillary}, VIPER~\cite{Richter_2017}, WildDash~\cite{Zendel_2018_ECCV} and KITTI~\cite{Menze2015CVPR}. These six diverse datasets contain images that range from congested urban city driving scenarios, rural areas, and highways to simulated scenes as well as indoor environments.

{\parskip=5pt
\noindent\textbf{Microsoft COCO:} There are 118K, 5K, and 20K images for training,  validation, and testing, respectively. The COCO dataset consists of 80 ‘thing‘ classes and 53 ‘stuff‘ classes. The images are of different resolutions.}

{\parskip=5pt
\noindent\textbf{Cityscapes:} The Cityscapes dataset consists of 2975, 500, and 1525 urban road scene images for training, validation, and testing, respectively. The images are a resolution of $2048 \times 1024$ pixels and contain 8 ‘thing’ classes and 11 ‘stuff’ classes.}

{\parskip=5pt
\noindent\textbf{Mapillary Vistas:} The Mapillary Vistas dataset is a large-scale driving dataset consisting of 18K, 2K, and 5K images for training, validation and testing, respectively. The images in this dataset have varying resolutions and range from $768 \times 1024$ pixels to $6000 \times 4000$ pixels. It contains 37 ‘thing’ classes and 28 ‘stuff’ classes.}

{\parskip=5pt
\noindent\textbf{VIPER:} The VIPER dataset consists of realistic imagery of virtual urban scenes, which are at a resolution of $1920 \times 1080$ pixels. It contains 18K, 2K, and 5K images for training, validation and testing, respectively. It comprises of 10 ‘thing’ classes and 13 ‘stuff’ classes.}

{\parskip=5pt
\noindent\textbf{WildDash:} The WildDash dataset is a collection of road scene images taken from all over the world with many diverse and challenging scenarios. The images were captured at a resolution of $1920\times1080$ pixels. It consists of 4256 training images and 776 test images. WildDash contains 13 ‘thing’ classes and 12 ‘stuff’ classes. We further split the training images randomly with a ratio of 80:20 to obtain the training and validation sets.}

{\parskip=5pt
\noindent\textbf{KITTI:} The KITTI dataset consists of 200 training and testing images where the resolution of the images are $375 \times 1242$ pixels and contains 8 ‘thing’ classes and 11 ‘stuff’ classes.}

\subsection{Data Augmentation}
\label{sec:augmentation}

The major problems while combining different benchmark datasets are various annotations schemes, difference in environments across datasets such as indoors and outdoors, and imbalance in the number of images available for training across different datasets. The difference in annotation schemes result in the same class being classified as being from the ‘stuff’ category in some of the datasets and being from the ‘thing’ category in others. For example, the pole class belongs to the ‘stuff’ category in the Cityscapes dataset but it belongs to the ‘thing’ category in the Mapillary Vistas dataset. To address this problem we create a unified label mapping space where all such objects are treated uniquely to address this discrepancy. In the aforementioned example, we treat the pole class in the Cityscapes dataset as a different class than the pole class in the Mapillary Vistas dataset. Therefore, our joint dataset has 109 ‘thing’ classes and 77 ‘stuff’ classes. To address the second problem that can lead to under or over-representation of a particular dataset due to the sheer number of training images, we replicate the dataset with fewer training examples for balanced training. In our case, the COCO dataset has 118K images which is roughly three times larger than all of the other datasets combined. Therefore, we replicate all the other dataset exclusive of COCO, three times and then combine it with COCO to form one large dataset or epoch.

Furthermore, we limit the longest side of the resolution of Mapillary Vistas to 1920 due to GPU memory constraints. We perform a limited set of random data augmentations including flipping and scaling within the range of $[0.5, 2.0]$. We also use variable crops to train our network such that for a given input image, it's original size i.e void of any scaling augmentation is considered as the crop size for the image. We also perform training on crops due to memory constraints.

\subsection{Training Protocol}
\label{sec:protocol}

We initialize the backbone of our EffPS\_b1bs4\_RVC architecture with weights from the EfficientNet model pre-trained on the ImageNet dataset and initialize the weights of the iABN sync layers to 1. We use Xavier initialization \cite{glorot2010understanding} for the other layers, zero constant initialization for the biases and we use Leaky ReLU with a slope of 0.01. We use the same hyperparameters as \cite{mohan2020efficientps} unless explicitly mentioned in this report. We train our model with Stochastic Gradient Descent (SGD) with a momentum of $0.9$ using a multi-step learning rate schedule i.e. we start with an initial base learning rate and train the model for certain number of iterations, followed by lowering the learning rate by a factor of 10 at each milestone and continue training for 10 epochs. We use an initial learning rate $lr_{base}$ of 0.01 and successively reduce it by a factor of 10 at 400k and 520k iterations. At the beginning of the training, we have a warm-up phase where the $lr_{base}$ is increased linearly from $\frac{1}{3}\cdot lr_{base}$ to $lr_{base}$ in 200 iterations. We train our EffPS\_b1bs4\_RVC with a batch size of 4 on 4 NVIDIA TITAN RTX GPUs where each GPU tends to a single-image. Please note that our benchmarked model is only trained on the training set i.e exclusive of the validation set.

\section{Testing}

We use test time augmentation such as flip and multi-scales only for COCO and Cityscapes. The scales that we use for COCO are [$0.75, 1.0, 1.25, 1.5, 1.75, 2.0$] and the scales for Cityscapes are [$0.75, 1.0, 1.25, 1.5$]. For Mapillary Vistas dataset, we upsample the predictions with the longest side that we limited to 1920, to the original image size for benchmark submission. The $min_{sa}$ parameter used in the panoptic fusion module is set to $512$ for COCO and $2048$ for the other datasets.

\section{Benchmark Results}

\begin{figure}
\centering
\footnotesize
\setlength{\tabcolsep}{0.1cm}
{\renewcommand{\arraystretch}{0.5}
\begin{tabular}{p{0.5cm}p{5.5cm} p{5.5cm}}
& \multicolumn{1}{c}{Image} & \multicolumn{1}{c}{Panoptic Prediction} \\
\\
\rot{(a) COCO} & \raisebox{-0.4\height}{\includegraphics[width=\linewidth]{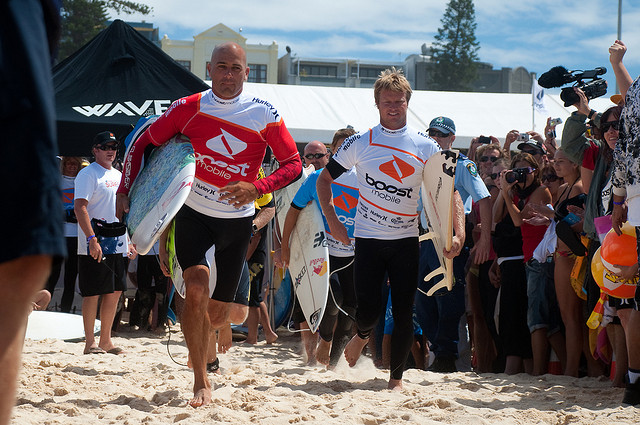}} & 
\raisebox{-0.4\height}{\includegraphics[width=\linewidth]{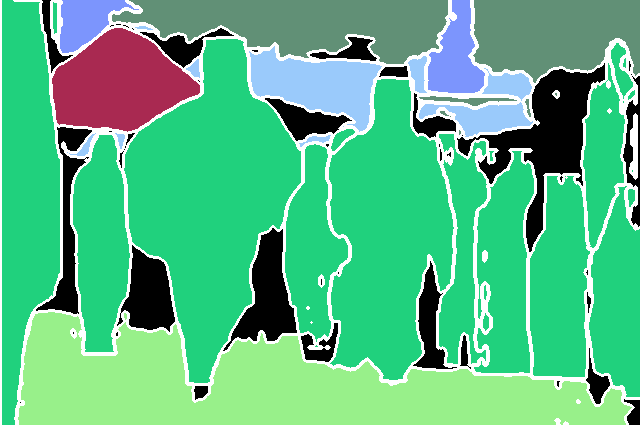}} \\
\\
\rot{(b) Cityscapes} & \raisebox{-0.4\height}{\includegraphics[width=\linewidth]{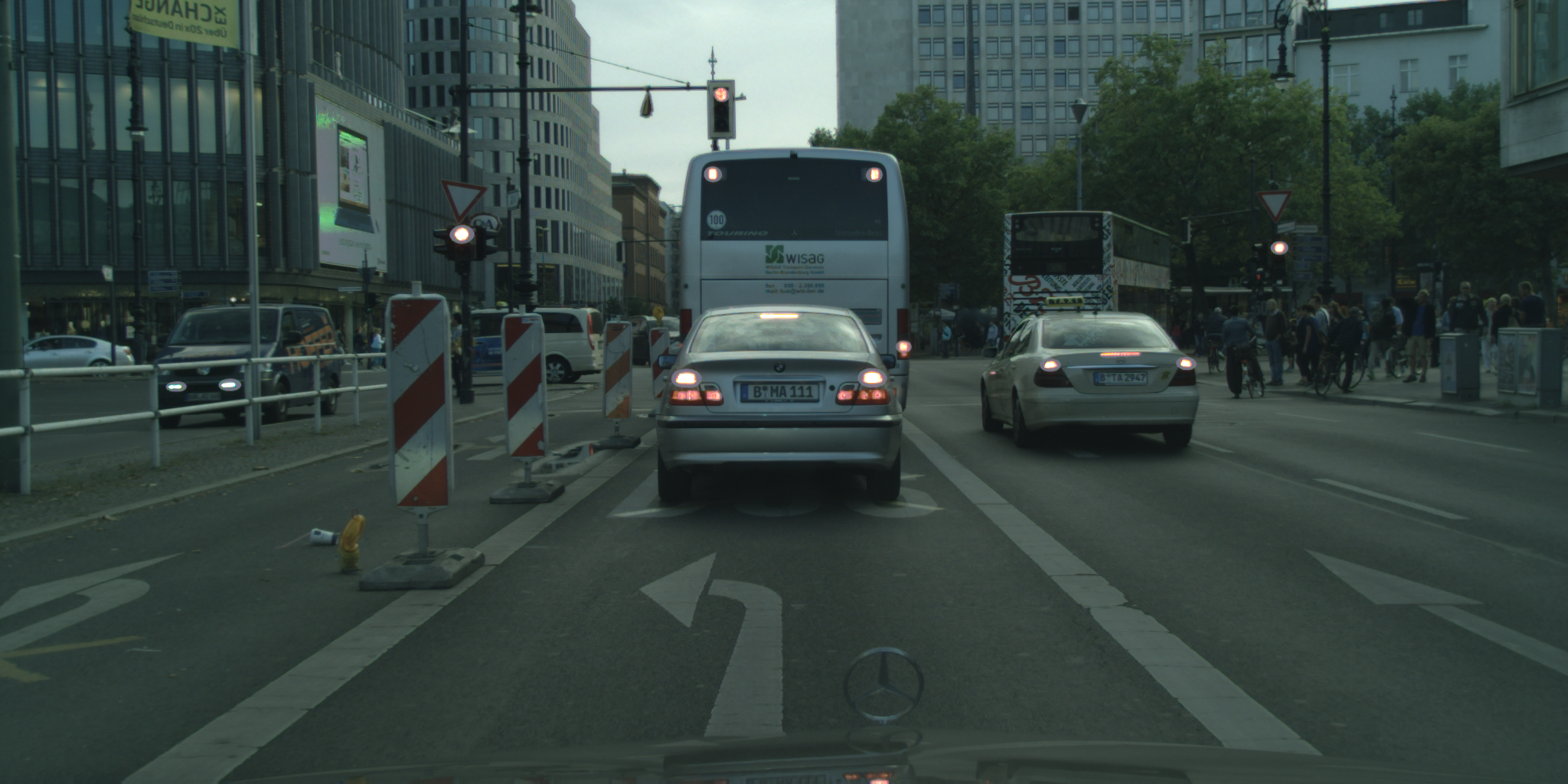}} & 
\raisebox{-0.4\height}{\includegraphics[width=\linewidth]{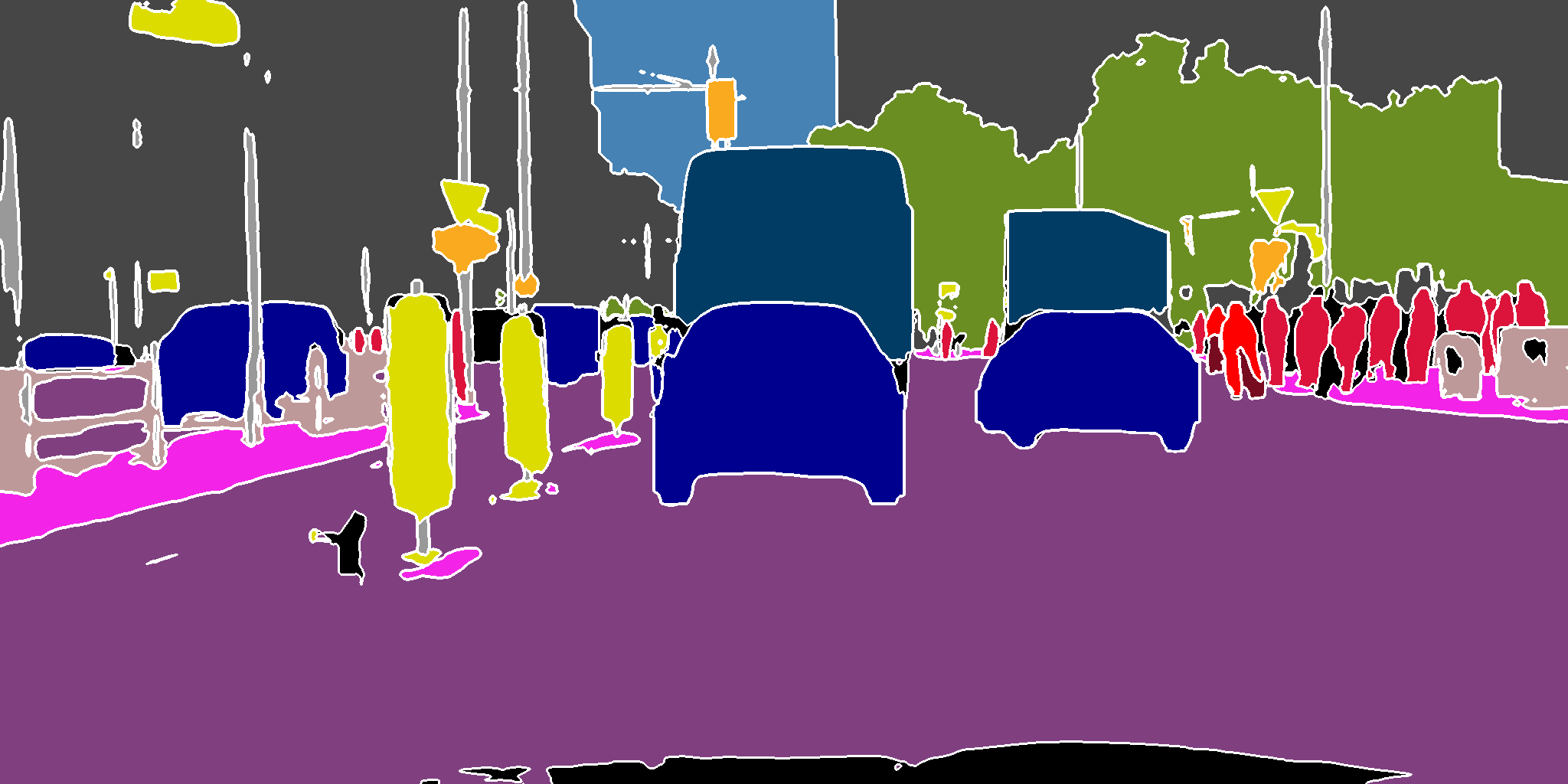}} \\
\\
\rot{(c) Mapillary Vistas} & \raisebox{-0.4\height}{\includegraphics[width=\linewidth]{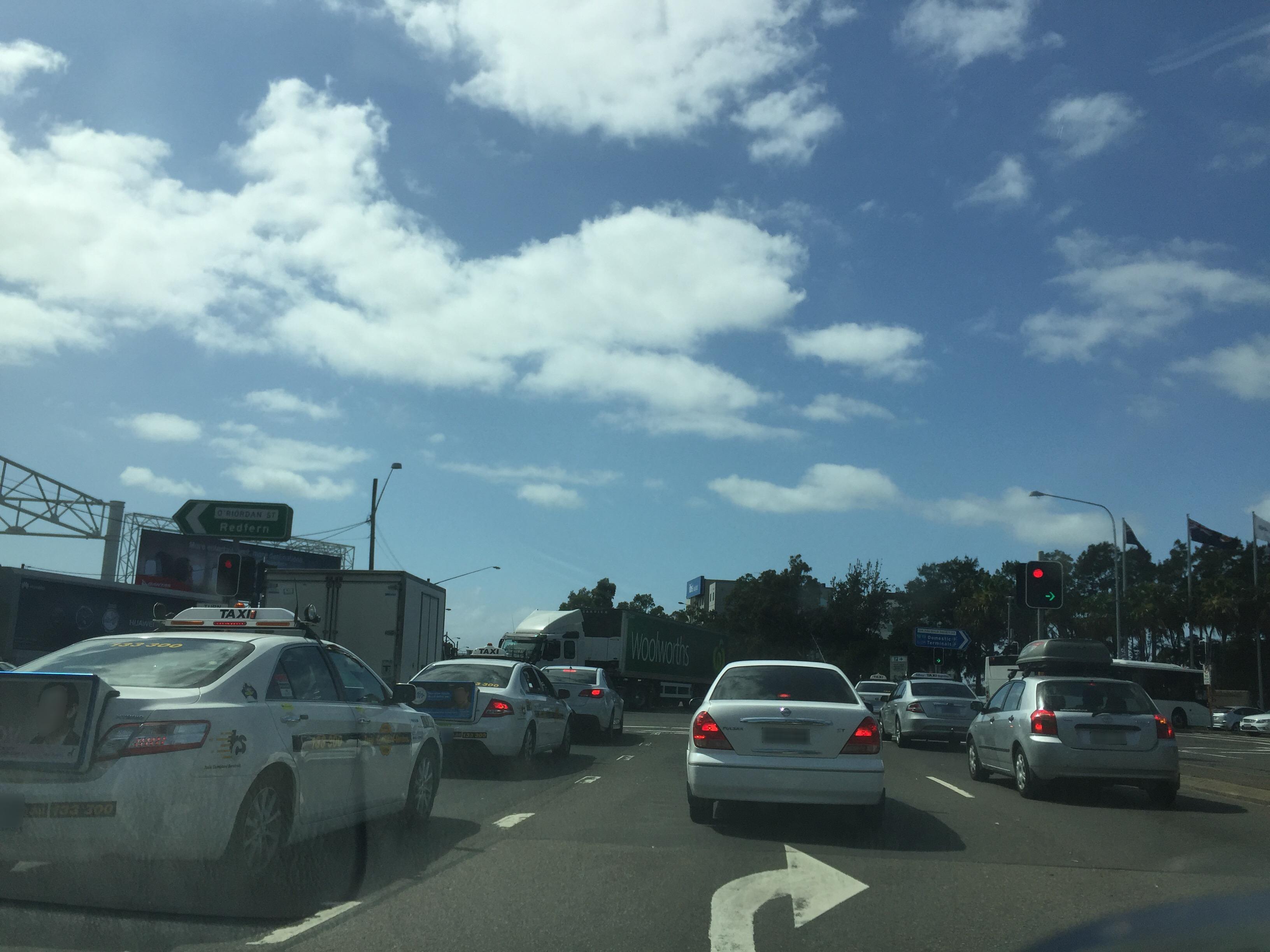}} &
\raisebox{-0.4\height}{\includegraphics[width=\linewidth]{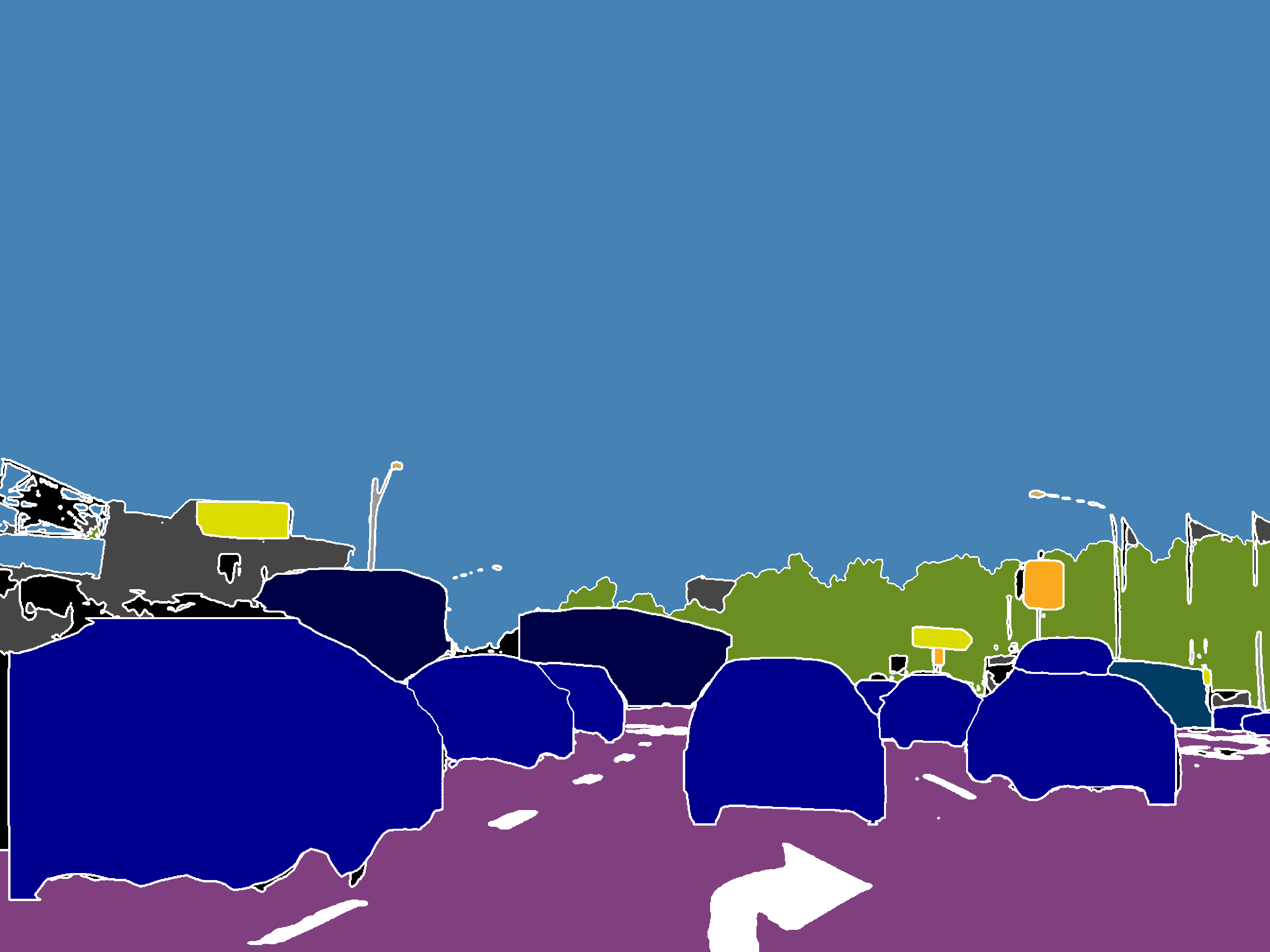}} \\
\\
\rot{(d) VIPER} & \raisebox{-0.4\height}{\includegraphics[width=\linewidth]{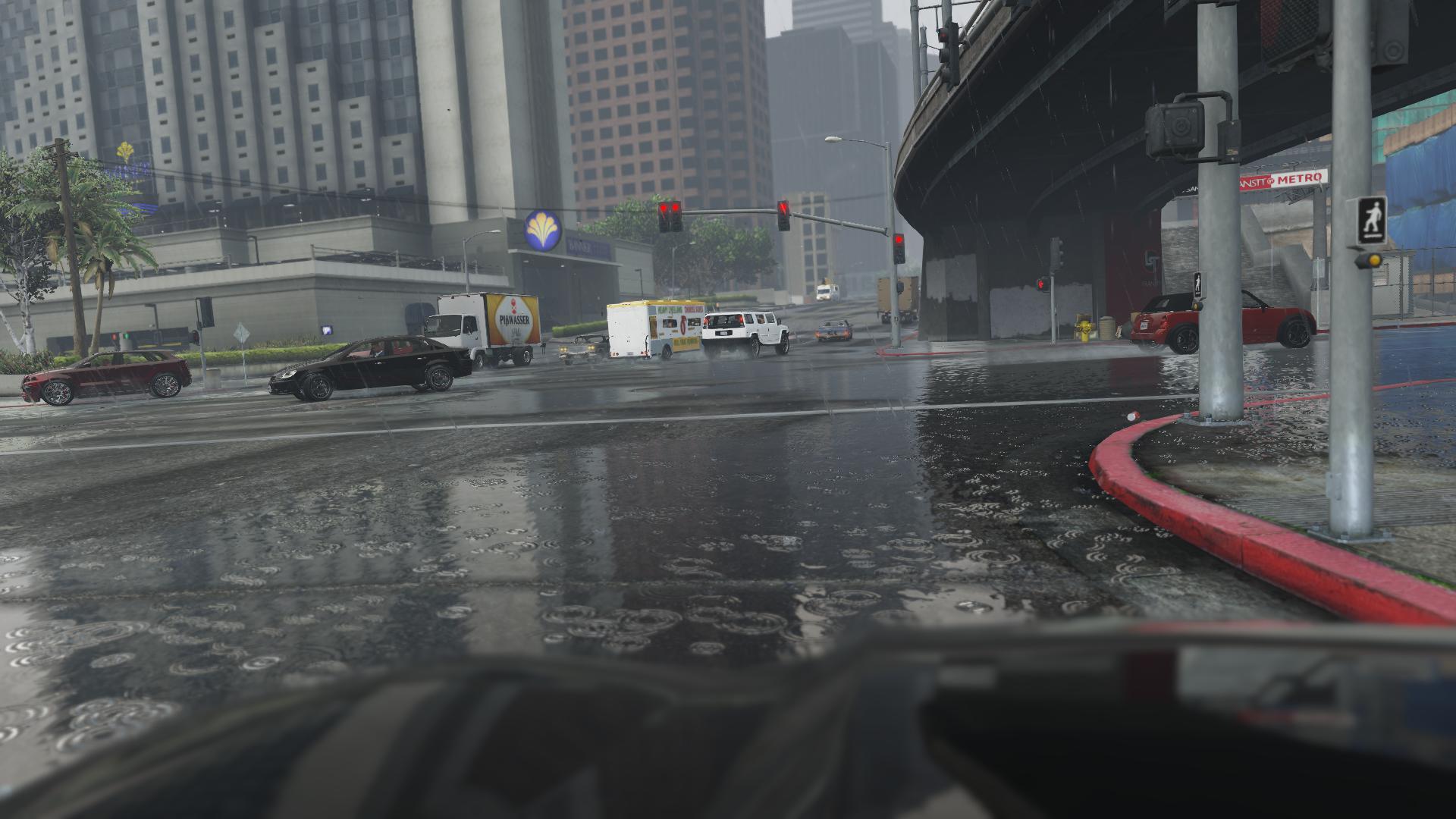}} &
\raisebox{-0.4\height}{\includegraphics[width=\linewidth]{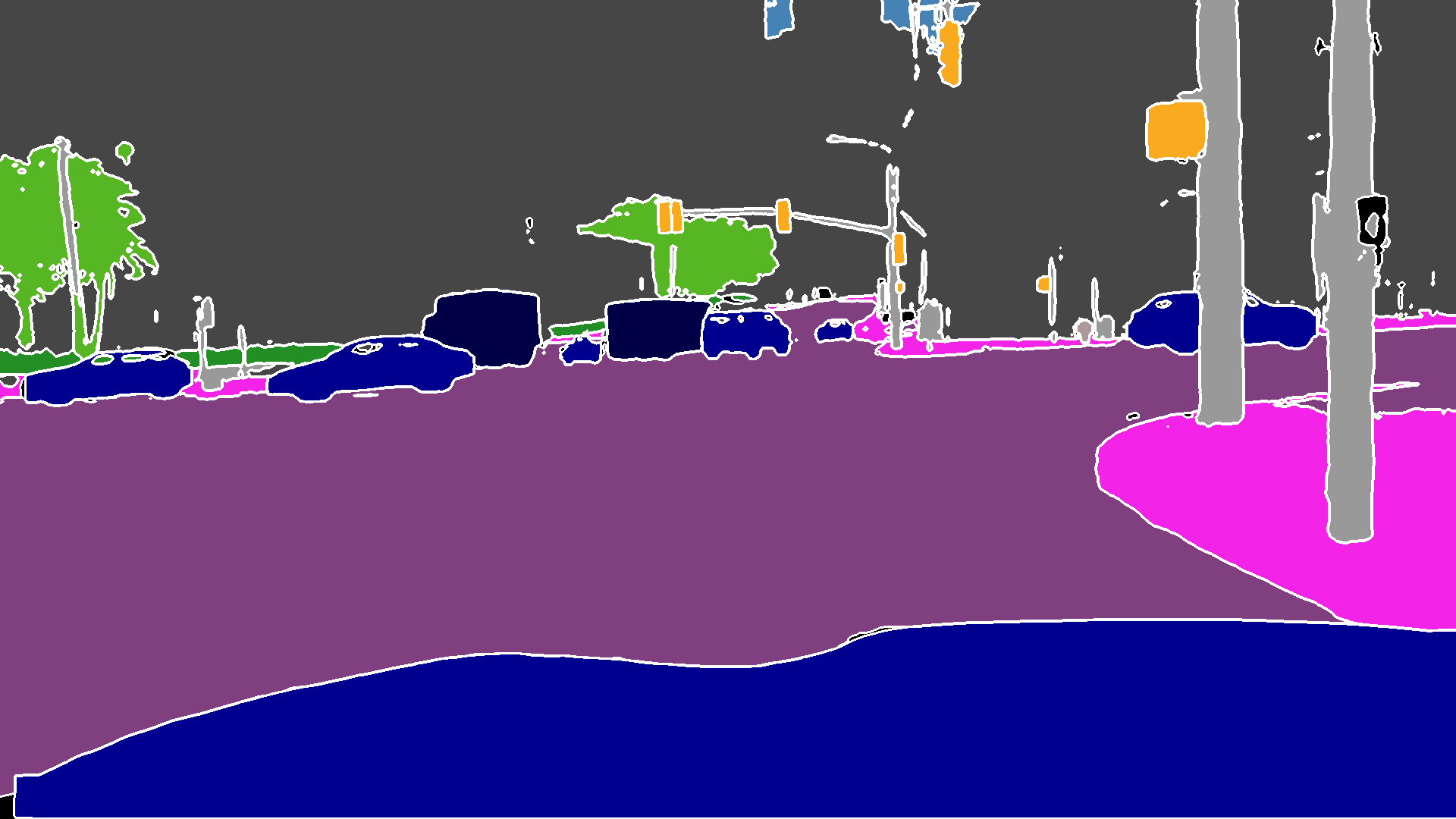}} \\
\\
\rot{(e) WildDash} & \raisebox{-0.4\height}{\includegraphics[width=\linewidth]{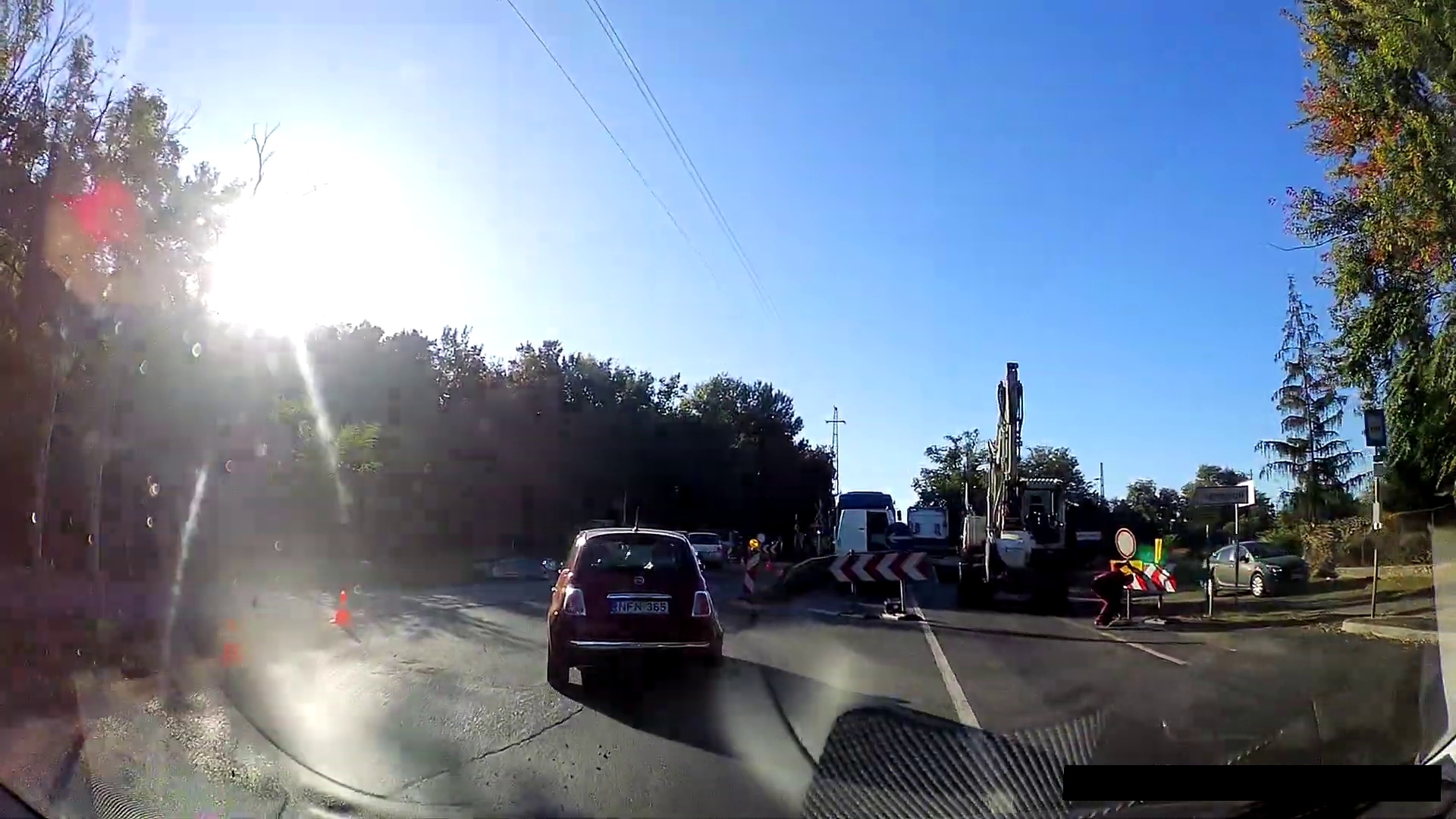}} &
\raisebox{-0.4\height}{\includegraphics[width=\linewidth]{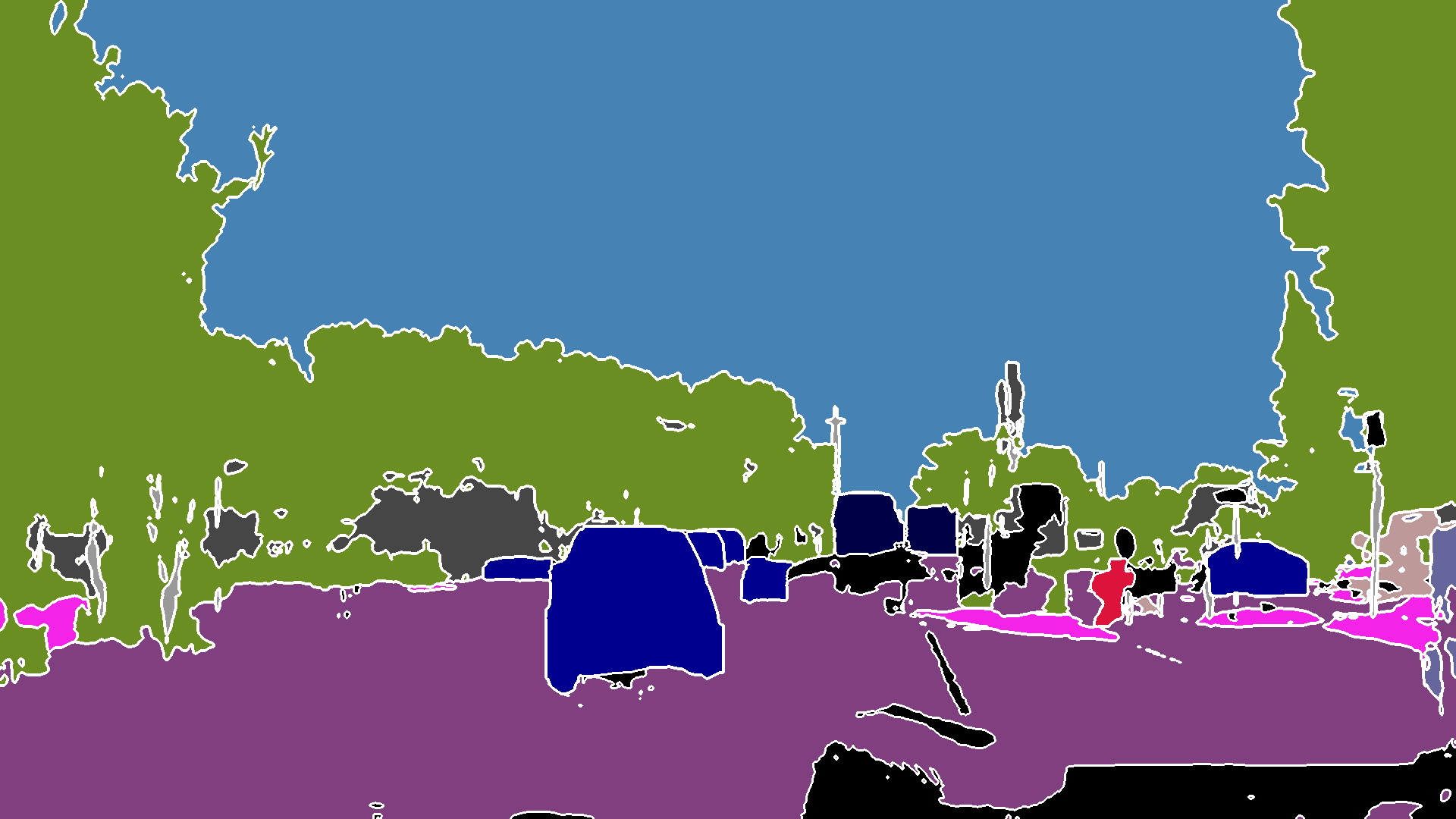}} \\
\end{tabular}}
\caption{Example images from the Robust Vision Challenge (RVC) 2020 Panoptic Segmentation Benchmarks namely, COCO, Cityscapes, Mapillary Vistas, VIPER and WildDash. The images exhibit wide variations in scenes ranging such as rainy weather condition, optical glare, cluttered urban scenes with occluded objects and variations in camera perspective.}
\label{fig:all}
\end{figure}

\tabref{tab:RVCBenchmark} presents the panoptic segmentation benchmark results of EffPS\_b1bs4\_RVC on the five datasets from the Robust Vision Challenge 2020. For simplicity, we only report the PQ, SQ and RQ scores. Other metric values can be accessed via the corresponding benchmark servers. Furthermore, \figref{fig:all} shows example results from each of the datasets in the RVC 2020 challenge.

\begin{table*}
\centering
\caption{Robust Vision Challenge (RVC) 2020 Panoptic Segmentation Benchmarks. PQ: Panoptic Quality, SQ: Segmentation Quality, RQ: Recognition Quality and MVD: Mapillary Vistas Dataset.}
\label{tab:RVCBenchmark}
\begin{tabular}{p{2.5cm}p{1.5cm}p{1.5cm}p{1.5cm}}
\toprule
Dataset & PQ & SQ  & RQ  \\
 &  $(\%)$ & $(\%)$ & $(\%)$ \\
\noalign{\smallskip}\hline\hline\noalign{\smallskip}
COCO & $20.0$ & $70.0$ & $25.0$  \\
Cityscapes & $48.0$ & $78.3$ & $59.2$  \\
MVD & $18.7$ & $59.0$ & $24.5$  \\
VIPER & $28.6$ & $72.1$ & $35.8$  \\
WildDash & $28.2$ & $31.8$ & $36.4$  \\
\bottomrule
\end{tabular}
\end{table*}

\section{Conclusions}

In this report, we presented our EffPS\_b1bs4\_RVC architecture that achieves the first place in the Robust Vision Challenge 2020 for the Panoptic Segmentation task. The challenge has a high level of difficulty than any other benchmark since it measures the performance of a solution on the combination of different challenging datasets that have vastly varying characteristics but at the same time exhibits the robustness of the given solution. Consequently, our result in RVC 2020 shows the robustness of EfficientPS across various benchmarks, as our winning model is essentially a lightweight version of it. Additionally, the performance of our model can be considerably improved by training for more epochs with higher batch sizes and with the inclusion of validation set in the training set.

%
%
\nocite{valada2016convoluted, valada2016towards, valada2018incorporating, boniardi2019robot}
\bibliographystyle{splncs04}
\bibliography{eccv2020submission}
\end{document}